\documentclass{llncs}
\usepackage{cite}
\usepackage{amsmath,amssymb,amsfonts}
\usepackage{graphicx}
\usepackage{textcomp}
\usepackage{xcolor}
\usepackage{balance}
\usepackage{subcaption}
\usepackage{enumitem}
\usepackage{rotating}
\usepackage{multirow}
\usepackage{algorithm,algpseudocode}
\usepackage[hyphens]{url}
\usepackage{newclude}
\usepackage{listings}
\usepackage{wrapfig}
\usepackage{comment}

\newcommand{\interop}{\texttt{inter\_op\_parallelism\_threads}}
\newcommand{\intraop}{\texttt{intra\_op\_parallelism\_threads}}
\newcommand{\batchsize}{\texttt{batch\_size}}
\newcommand{\kmpblocktime}{\texttt{KMP\_BLOCKTIME}}
\newcommand{\ompnumthreads}{\texttt{OMP\_NUM\_THREADS}}

\begin{document}

\title{Automatic Tuning of Tensorflow’s CPU Backend using
 Gradient-Free Optimization Algorithms\thanks{Accepted for publication in Machine Learning on HPC
 Systems workshop (MLHPCS) held along with International Supercomputing
Conference (ISC), July 2, 2021.}}



\newcommand{\printfnsymbol}[1]{%
  \textsuperscript{\@{#1}}%
}

\renewcommand{\thefootnote}{\fnsymbol{footnote}}

\author{Derssie Mebratu\inst{1} \protect\footnote[5]{equal contribution} \and
        Niranjan Hasabnis\inst{2} \printfnsymbol{$\ddag$} \and
        Pietro Mercati\inst{2} \and
        Gaurit Sharma\inst{1} \and
        Shamima Najnin\inst{1}}

\institute{Intel Corporation, Hillsboro, Oregon, USA
\and Intel Labs, Santa Clara, California, USA \\
\email{\{derssie.d.mebratu, niranjan.hasabnis\}@intel.com} \\
\email{\{pietro.mercati, gaurit.sharma, shamima.najnin\}@intel.com}
}

\maketitle

\begin{abstract}

Modern deep learning (DL) applications are built using DL libraries and
frameworks such as TensorFlow and PyTorch. These frameworks have complex
parameters and tuning them to obtain good training and inference performance
is challenging for typical users, such as DL developers and data
scientists. Manual tuning requires deep knowledge of the
user-controllable parameters of DL frameworks as well as the underlying
hardware. It is a slow and tedious process, and it typically delivers
sub-optimal solutions.
In this paper, we treat the problem of tuning parameters of DL frameworks to
improve training and inference performance as a black-box optimization problem.
We then investigate applicability and effectiveness of Bayesian optimization
(BO), genetic algorithm (GA), and Nelder-Mead simplex (NMS) to tune the
parameters of TensorFlow’s CPU backend.  While prior work has already
investigated the use of Nelder-Mead simplex for a similar problem, it does not
provide insights into the applicability of other more popular algorithms.
Towards that end, we provide a systematic comparative analysis of all three
algorithms in tuning TensorFlow's CPU backend on a variety of DL models. Our
findings reveal that Bayesian optimization performs the best on the majority of
models. There are, however, cases where it does not deliver the best
results.

\end{abstract}


\section{Introduction}

In recent years, deep learning has gained significant momentum in academic
research as well as in production to solve real-world problems.  For example,
deep learning applications in the areas of speech recognition (e.g., Amazon
Alexa, Apple Siri, Google Assistant, etc.), language translation (e.g., Google
Translate), and recommendation systems (e.g., Netflix movie recommendations,
Amazon product recommendations, etc.) are already part of everyday life.
Interest in deep learning is fueled by the vast availability of both open-source
and proprietary data as well as by the continuous development of heterogeneous
computing platforms (e.g., CPU, GPU, TPU, etc.) and cloud resources (e.g.,
Amazon AWS, Google Cloud, Microsoft Azure, etc.) to process that data.

The availability of open-source deep learning software frameworks, such as
PyTorch~\cite{paszke:2019:pytorch} and TensorFlow~\cite{abadi:2016:tensorflow},
along with the suites of neural network models~\cite{tensorflow-model-zoo}
enables fast deployment of deep learning models. Although deep learning
frameworks are relatively new software systems, they essentially employ software
designs that are similar to other existing software systems, particularly
compilers.  Deep learning frameworks accept models written in high-level
languages such as Python.  Similar to the compilers for high-level languages,
these models are either interpreted directly (as for Python itself) or converted
(``lowered'', in compilers parlance, such as for C and C++ languages) into a
low-level data-flow graph that is later executed.  Before the models are
executed, the framework runtime schedules the computations from their data-flow
graphs onto the backends for the hardware
devices~\cite{abadi:2016:tensorflow}.  Consequently, the training or inference
performance of a deep learning model partly depends upon runtime's scheduling
decisions.

TensorFlow's default CPU backend is implemented using an open-source library
named Eigen~\cite{eigen}. TensorFlow's Eigen CPU backend enables efficient
execution on multicore CPUs by offering a configurable threading model to
exploit the concurrency that is typically present in TensorFlow's data-flow
graphs.  Specifically, vertices in TensorFlow's data-flow graphs represent
computations, and they can have data dependencies (i.e., input edges from other
vertices) and control dependencies (i.e., a scheduling constraint specified by
the user or TensorFlow framework). The Eigen CPU backend relies on PThreads
library for multi-threading, and its threading model offers two configurable
parameters: \emph{(i)} {\interop}: the maximum number of independent
computations to execute in parallel, and \emph{(ii)} {\intraop}: the maximum
number of threads to use for executing a single computation.  Unfortunately, the
Eigen CPU backend has shown sub-optimal performance on several Intel Xeon CPU
platforms~\cite{hasabnis:2018:tensortuner,
Ould-Ahmed-Vall:2017:intel-tensorflow-paper}.  Consequently, Intel contributes
with its own CPU backend to TensorFlow~\cite{intel-tensorflow-install-webpage}
that delivers orders of magnitude of performance improvement over the Eigen CPU
backend.  Intel's CPU backend uses the OpenMP~\cite{openmp} library for
multi-threading and adds another configurable parameter to
TensorFlow's threading model: \emph{(iii)} {\ompnumthreads}: the number of
OpenMP threads to use for executing a single computation with Intel's backend.

TensorFlow's configurable CPU threading model enables end-users to improve
performance of their models by tuning TensorFlow to the target hardware. As
mentioned in TensorFlow's ``Binary Configuration'' section
~\cite{tensorflow-perf-tuning}, a savvy user can tune the model by finding
values of {\interop} and {\intraop} by ``finding the right configuration for
their specific workload and environment''. The guide, however, does not discuss
how to find the right configuration in practice.  Unfortunately, it is
unrealistic to expect that a deep learning application developer or a data
scientist would know the optimal parameter configurations as these
configurations are intimately related to detailed knowledge of the framework and
the underlying hardware.  Considering this limitation, Intel provides specific
configurations~\cite{intel-tensorflow-perf-webpage} for popular deep learning
models, such as ResNet50, on commonly-used Intel Xeon platforms, such as the
latest generation Intel Xeon CPUs (codenamed IceLake).  However, any deviation
from this standard setup, for example with a new model or a new hardware
platform, could mean that the provided settings may not deliver the optimal
performance.  The alternative of relying on the default values of the parameters
can be acceptable in some situations such as early prototyping.  However, they
usually deliver sub-optimal performance~\cite{hasabnis:2018:tensortuner}.

A common approach to search for the optimal configuration is manual search. Manual
search, however, is a tedious activity, leading to sampling only a few
configurations, and it also relies on the expertise of the user.  A naive
approach of exhaustive search is feasible for small search spaces, but
becomes unfeasible as the search time grows exponentially in the number
of parameters.  For instance, in our experiments, the exhaustive search run
for the optimal configuration of TensorFlow's threading model for
ResNet50 inference took close to a month of CPU time on a multi-core Intel
Xeon platform. The search space consisted of roughly 50000 points. While this
could be acceptable in research and development settings, it would not be acceptable
in production environments.

Hasabnis~\cite{hasabnis:2018:tensortuner} offers an excellent description of the
problem and an auto-tuning solution, called TensorTuner, to configure
TensorFlow's threading model for CPU backend.  Specifically, TensorTuner uses a
black-box optimization algorithm named Nelder-Mead simplex. This solution
addresses both the issues described earlier: obtaining the best performance and
systematically configuring the parameters of the threading model.  Nelder-Mead
simplex, however, is a local optimization algorithm. And popular global
optimization algorithms, such as Bayesian optimization and genetic algorithm,
have been demonstrated to perform successfully in system tuning
tasks~\cite{Li:2021:rambo}.  Hasabnis does not consider these alternatives and
leaves the open question of the best algorithm for this problem.

In this paper, we analyze the effectiveness of Bayesian optimization, genetic
algorithm, and Nelder-Mead simplex to tune TensorFlow's threading model for
various deep learning models. Unlike TensorTuner that focuses solely on deep learning
models from image recognition domain, we consider models belonging to a variety
of domains. Furthermore, we also consider a larger set of performance-sensitive
parameters by considering {\batchsize} and {\kmpblocktime}, which are not
considered by TensorTuner. Finally, we perform detailed comparative analysis of
the performance of all three algorithms and discuss our findings.

\subsection{Contributions}

In this paper we make following contributions:
\begin{enumerate}

\item We evaluate and compare the effectiveness of Bayesian optimization,
genetic algorithm, and Nelder-Mead simplex to automatically tune
performance-critical parameters of TensorFlow's Intel-CPU backend.

\item We consider several deep learning models, written for a variety of use cases
such as image recognition, language translation, etc. Prior
work for this problem has focused on the models used in image recognition only.

\item We analyze performance of each optimization algorithm. The
analysis provides us insights on the performance behavior of the algorithms
and the classes of problems for which they could be more successful.

\end{enumerate}

This paper is organized as follows. Section~\ref{section:background} provides
the necessary background information about TensorFlow and black-box
optimization algorithms.  Section~\ref{section:methodology} and
Section~\ref{section:evaluation} present our evaluation methodology and results,
respectively.  Section~\ref{section:related_work} presents the related
work, while Section~\ref{section:conclusion} concludes the paper.

\section{Background}
\label{section:background}

This section provides a brief introduction to TensorFlow, Bayesian optimization
(BO), genetic algorithm (GA), and Nelder-Mead simplex (NMS) algorithm. The
description is not meant to be exhaustive, but sufficient to understand the
results presented in the experimental section.

\subsection{TensorFlow}

TensorFlow~\cite{abadi:2016:tensorflow}, initially released in 2015, is an open-source
library for machine learning and deep learning that is developed by the Google
brain team. It is a multi-system library that supports Linux, MacOS and
Windows and is implemented primarily in Python, C++, and CUDA.

TensorFlow supports machine learning and deep Learning models implemented in
languages such as Python and Javascript. It also offers high-level Keras APIs to
enable quick model development and prototyping. It supports execution on various
hardware devices such as CPUs, GPUs, TPUs, etc., and provides a variety of
tools (e.g., TensorFlow eXtended, TensorFlow Lite, TensorFlow.js, etc.)
that enable easy deployment of trained models on those devices.

\paragraph{Execution modes.} TensorFlow's current version (version 2) supports
two modes of execution: eager mode and graph mode. The eager mode is similar to
Python's interpreter mode in that Keras/Python APIs invoked by programmers are
interpreted immediately. The graph mode, on the other hand, leverages the
concept of \emph{lazy evaluation} and builds an intermediate data-flow graph
representation before executing it. Although the eager mode enables faster
prototyping and model development, its performance is typically lower than the
graph mode. This is because the graph mode can perform global optimizations over
the data-flow graphs that are not possible in the eager mode.

\begin{figure*}[!t]
\begin{minipage}[c]{.45\textwidth}
\begin{footnotesize}
\begin{verbatim}
import tensorflow as tf
from tf.keras import Input, Model
from tf.keras.layers import Dense
from tf.nn import relu

in = Input(shape=(3,),
               batch_size=2)
out = Dense(4,
            use_bias=True,
            activation=relu)(in)
model = Model(in, out)
\end{verbatim}
\end{footnotesize}
\caption{Python implementation of the model for $y$ = $W.x + b$}
\end{minipage}
\hfill
\begin{minipage}[c]{.5\textwidth}
\includegraphics[width=\linewidth]{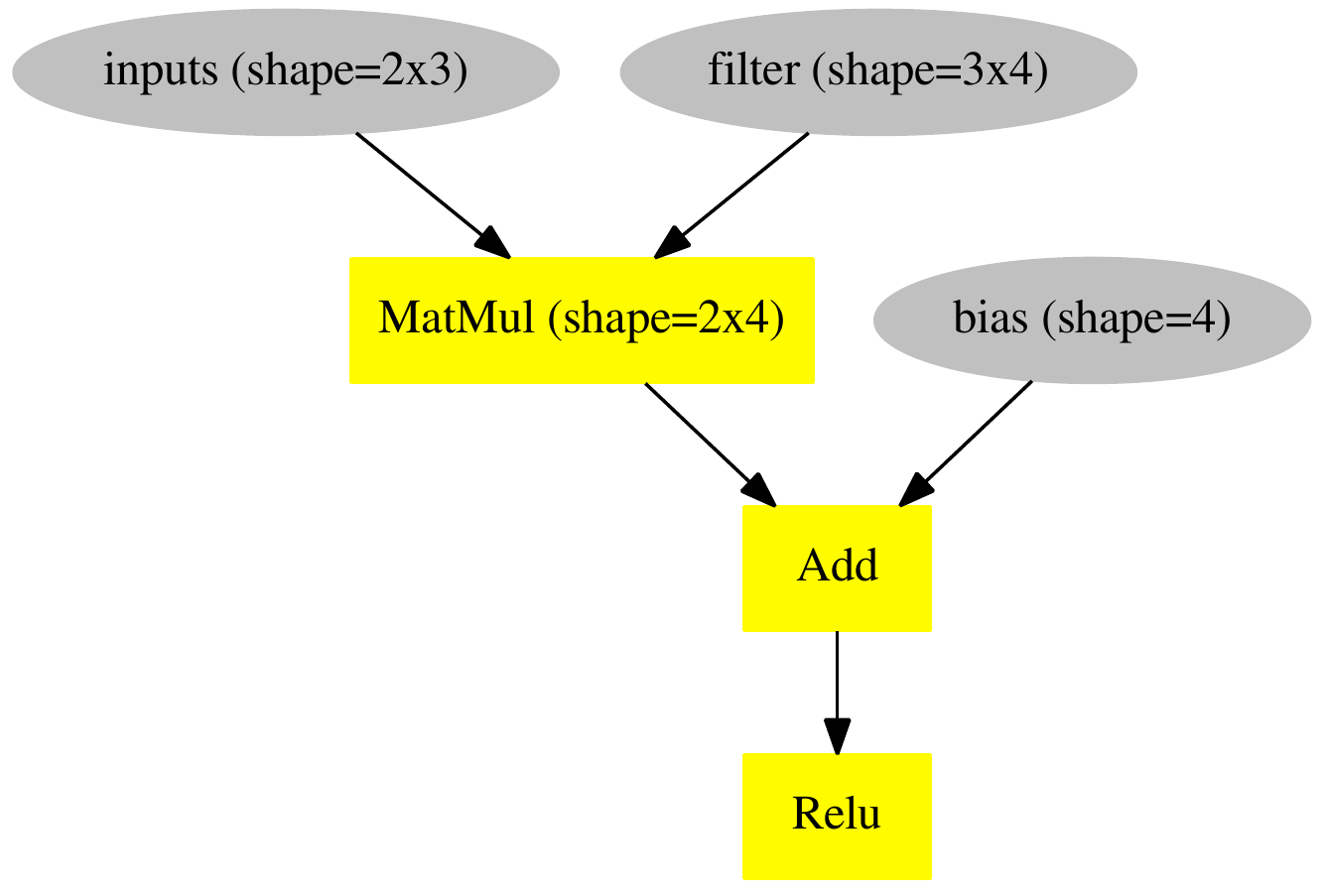}
\caption{Data-flow graph for the model (yellow boxes show computations, while
gray boxes show variables/tensors)}
\end{minipage}

\caption{TensorFlow model for $y$ = $W.x + b$ (left) and its data-flow graph (right)}
\label{fig:tf_dataflow_graph}
\end{figure*}

\paragraph{Tensors and data-flow graph.} TensorFlow's data-flow graph represents
high-level machine learning models by representing computations as vertices and
inputs/outputs of the computations as incoming and outgoing edges. In
TensorFlow, inputs and outputs of the computations are called \emph{tensors}.
This name, borrowed from algebra, indicates objects that can be represented as
N-dimensional matrices.  Edges in the data-flow graph carry tensors between the
computations, and thus the name ``data-flow'' graph.
Figure~\ref{fig:tf_dataflow_graph} shows a Python implementation of the model
for ($y$ = $W.x + b$), a basic operation of a neural network, written using
TensorFlow's Keras APIs. The figure also shows its corresponding data-flow
graph. The arrows in the figure represent tensors, and the directions of the arrows
correspond to the directions of the data flow.

\paragraph{Scheduling and execution.} Edges in TensorFlow's data-flow graph can
represent two different types of dependencies: \emph{data dependencies} and
\emph{control dependencies}. An edge that feeds an output of an operation $X$ to
an operation $Y$ constitutes a data dependency of the operation $Y$ on the
operation $X$.  It means that the operation $Y$ cannot be executed or
scheduled for execution until the operation $X$ has finished its execution.
Control dependencies, on the other hand, represent scheduling
constraints between the operations and can be inserted (by the user or
automatically) to enforce a particular scheduling order.  As an example, solid
arrows in Figure~\ref{fig:tf_dataflow_graph} only show data dependencies.  There
are no control dependencies in the figure.

\paragraph{CPU threading model.} Data and control dependencies in TensorFlow's
data-flow graph help in enforcing a particular execution order of the
operations. However, the execution order can still be partial.  In particular,
operations in the data-flow graph that do not have any direct or indirect
dependencies can be executed in any order. With reference to
Figure~\ref{fig:tf_dataflow_graph}, variables \texttt{inputs} and
\texttt{filter} of \texttt{MatMul} can be accessed either concurrently or
serially (\texttt{inputs} after \texttt{filter} or vice versa.) TensorFlow's
threading model for multi-core CPU devices allows users to exploit this
\emph{concurrency} in the data-flow graphs by setting two parameters of the
threading model: {\interop} and {\intraop}. {\interop} specifies the maximum
number of operations that can be executed concurrently, while {\intraop}
specifies the maximum number of threads that can be used by a single operation.
Both parameters together restrict the total number of threads that will be used
to execute a data-flow graph.  Therefore, they are used to control
over-subscription of a device. The default values of these parameters are
decided by TensorFlow's runtime depending on the underlying platform. The
default values, however, are not adapted further to the input machine
learning model, thus missing an opportunity for performance improvement.

\subsection{Black-box Optimization Algorithms}

Tuning the parameters of a software system to improve performance metrics, such
as execution time or memory consumption, can be formulated as a black-box
optimization problem. ``Black-box'' here refers to any system for which
analytical description or gradients is not available, and instead it is only
possible to query a configuration of the inputs and measure the corresponding
output.  Therefore, the problem cannot be solved with gradient-based techniques,
but requires gradient-free optimization algorithms of non-linear systems.  The
algorithms investigated in this work belong to the class of gradient-free
optimization algorithms, and all can solve this problem.  However, they have
fundamentally different behaviors as explained in the following sections.

Formally, a black-box has input $x \in X$, where $X$ is the
solution space. Note that $X$ has, in general, $d$ dimensions, so
$x = x_{0}, x_{1}, ..., x_{d-1}$. The system can be described by
an objective function $f$, which is unknown but measurable. A
measurement or evaluation corresponding to input $x$ would be $y = f(x)$.
The optimization problem can be formulated as:
\[ \min_{x \in X} f(x) \]

The solution to this problem would be an input configuration $x^* = argmin
f(x)$, also called minimizer, and the only information that can be leveraged to
find this solution is the history of past $n$ measurements $D = \{(x_i,
y_i)\}_{i=0}^{n-1}$.

Different classes of algorithms have been developed to address black-box
systems, most notably model-based, evolutionary and heuristic, and all of them
are iterative in nature.

Bayesian optimization (BO) is a model-based algorithm, meaning that it uses
system evaluations to construct a surrogate model of the optimization
objective, and leverages the knowledge of the model to guide the selection of
the next configuration to evaluate. It is called Bayesian, because the model
employed is probabilistic, often a Gaussian Process, which is fundamental to
trade global exploration in the regions of large uncertainty with
local exploitation around the best solutions observed.  This is different from
more traditional models such as linear regression or neural network regression
that only return a predicted value for each input.  Instead, a probabilistic
model returns both a prediction and an estimate of uncertainty for that
prediction. This information is leveraged by BO at each iteration to guide the
selection of candidate solutions. For this, predictions and uncertainties are
used to evaluate an acquisition function, which takes large values in the
vicinity of promising past measurements or in the regions with large uncertainty.
After the initial model is ready, usually trained with a few random evaluations,
BO starts a loop of iterations. First, it computes and maximizes the acquisition
function. The solution maximizing the acquisition function is selected as the
next configuration to evaluate. Second, this configuration is applied to the
system and evaluated. Finally, the measurement provides a new data point, which
is used to update the surrogate model.

In this work we use Gaussian Processes (GPs) as the surrogate models. The GP is
highly ``data-efficient'', thus it achieves good accuracy with a relatively
small number of training points, and can be customized to model different
classes of functions by changing its ``kernel'' function. Finally, GPs have
convenient analytical properties that allow to train them with a closed-form
approach.
For the acquisition function we adopt
``SMSego'', because it is fast to compute and delivers state-of-the-art
performance.  For each point in the solution space, this function accepts
the prediction and the uncertainty from the surrogate model and estimates how likely
they can extend the best evaluation observed so far.
SMSego has been shown to have performance comparable to other best acquisition
functions, which are harder to implement and require
approximations.

Genetic algorithm (GA) belongs to the family of evolutionary algorithms, which
draw inspiration from biological phenomena such as reproduction.  Instead of
building an internal model, at each iterations, GA relies upon a fitness
function to select two ``best parent configurations'' from the history
of the evaluated configurations. Then, the parent configurations are manipulated via
crossover and mutation operations to generate a ``child'' configuration.  GA
reflects the process of natural evolution, in which genes determining a better
adaptation are mixed together with occasional mutations, which leads to stronger
and healthier generations.  More formally, the GA would take the history as an
input and reorder the input-output pairs based on a certain fitness
function $g(x_i, y_i)$. Then, it would pick the inputs of the two fittest pairs,
called ``parents'', and generate a new input by copying part of the components
from the first parent and the other from the second parent. This operation is
called crossover. Then, it might also change one or more component to purely
random values. This is referred to as mutation.  Evolutionary algorithms are
broadly used for their ease of implementation and configuration.



Finally, Nelder-Mead simplex (NMS) is a direct search heuristic method that
uses evaluations to build a simplex object in the space of objective function.
The next configuration to evaluate is selected by manipulating the simplex via
reflection, expansion and contraction operations. While simple to implement and
intuitive, NMS has a tendency to get stuck in local optima.

\section{Optimization Framework and Methodology}
\label{section:methodology}

In this section, we describe our automated optimization methodology for black-box
optimization of TensorFlow's CPU backend.

\begin{figure*}[!t]
\centering
\includegraphics[width=\linewidth]{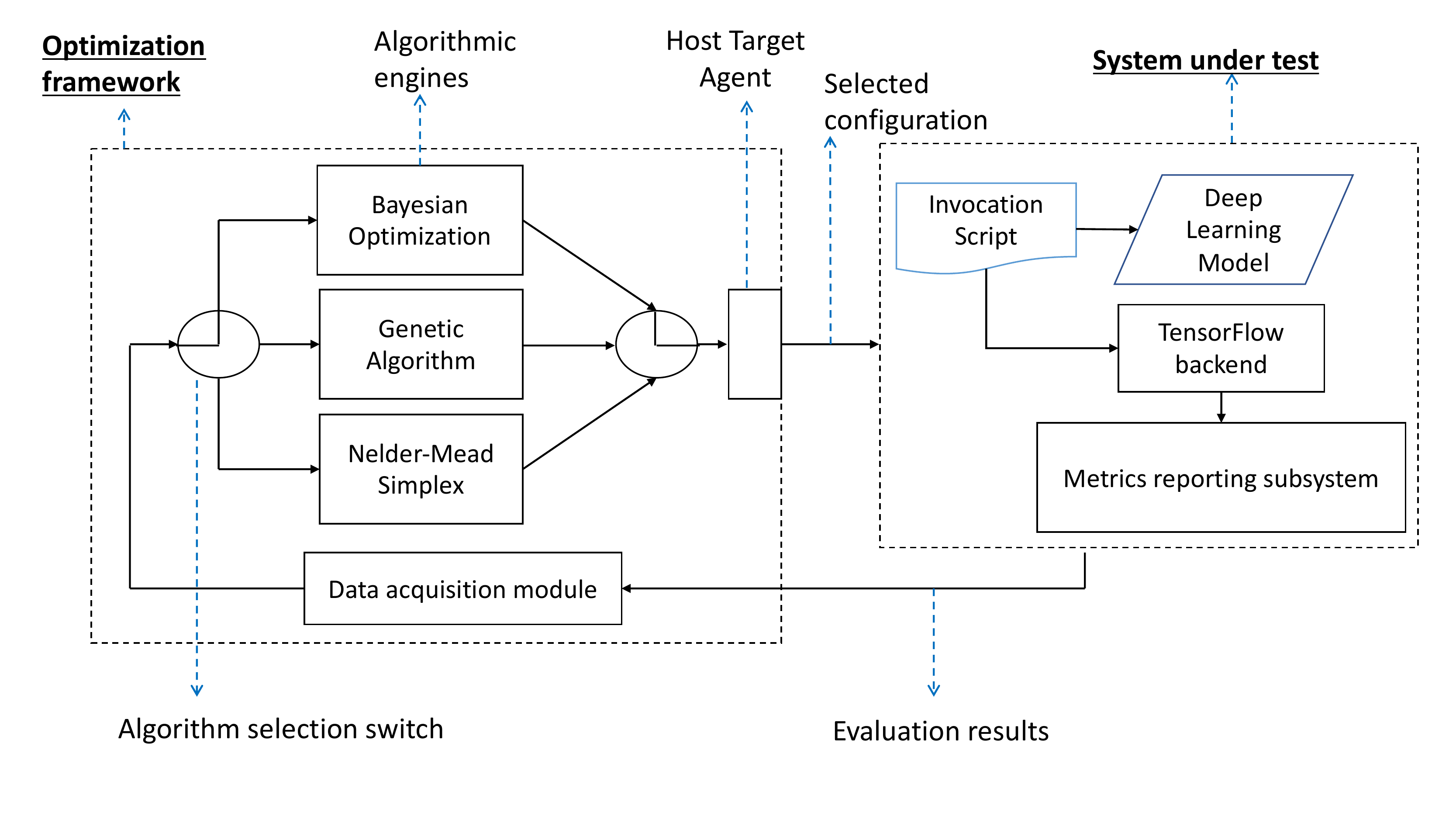}
\caption{Optimization framework and methodology}
\label{fig:methodology}
\end{figure*}

Figure~\ref{fig:methodology} shows the block diagram of the optimization setup.
The optimization framework (on the left) has different components, and it runs
on the host system. The algorithmic engines implement the black-box optimization
algorithms described in the previous section: Bayesian optimization, genetic
algorithm and Nelder-Mead simplex. The algorithm selection switch is configured
to exercise one engine at a time. This ensures that all engines can use the same
interface to TensorFlow for converting and applying the chosen parameters and
the same data acquisition module to retrieve evaluation results and update the
evaluation history.  On the right, the system under test is the target system,
and generically, it is any computing system that can execute TensorFlow models.
The only requirement on the system is that it allows applying the parameters and
measuring the corresponding output via some metric reporting subsystem (e.g. a
log file in the simplest case).  A clear separation of components ensures that
the workload performance is not affected by interference from the optimization
algorithm. It also enables us to run optimization algorithm on a relatively
less-powerful machine than the target machine.

The mapping of the optimization problem is realized as following. At each
iteration, an algorithmic engine selects a configuration $x$ of TensorFlow's
threading model parameters. Through the TensorFlow interface, the configuration
$x$ is converted into a command to set the values of parameters on the target
system.  Then, the optimization framework runs an evaluation on the target
system and evaluates the objective function $f(x)$ of the metric of interest,
such as images processed per second in the case of ResNet50.  The evaluation
provides a new data point, which is added to the global history of evaluations.
In case of Bayesian optimization, this is then used to retrain the Gaussian
process model, recompute the acquisition function and maximize to select the
next configuration.

\section{Evaluation}
\label{section:evaluation}

In this section, we discuss our evaluation of Genetic algorithm, Bayesian
optimization, and Nelder-Mead simplex algorithm to tune the inference throughput
of the selected deep learning models.  In other words, the objective function
was to maximize the throughput of performing inference over every model.

\subsection{Experimental Setup}

Before we present our results, we describe the evaluation setup.

\paragraph{\textbf{Hardware configurations:}}

We used a dual-socket, 22-core Intel Xeon E5-2699 v4 processor
(codenamed Broadwell) as the host system and a dual-socket, 24-core 2nd-generation
Intel Xeon Scalable Gold 6252 processor (codenamed Cascade Lake) as
the target system. The processor for the host system was configured to run at
3.6 GHz with 384 GB of physical memory, while the processor for the target
system was configured to run at 3.9 GHz with hyper-threading turned on and with
512 GB of physical memory.  Both the servers were running Ubuntu-18.04 operating
system.

\paragraph{\textbf{Software configurations:}} We installed Intel-optimized
TensorFlow v1.15~\cite{intel-tensorflow-install-webpage} on the target system to
run the deep learning models. This TensorFlow version uses version v0.20.6 of
Intel's Math Kernel Library for Deep Neural Networks (oneDNN). oneDNN is an
open-source, cross-platform high-performance library of basic building blocks
for deep learning applications~\cite{onednn}.
We used Python 3.7.7 to run the TensorFlow benchmarks.

\paragraph{\textbf{TensorFlow models:}} Intel
Model Zoo~\cite{intel-tensorflow-model-zoo} provides a suite of popular deep
learning models that are optimized for various versions of
Intel-optimized TensorFlow. We used SSD Mobilenet, ResNet50, Transformer-LT,
BERT, and NCF models from the Intel provided suite. We selected the models such
that they cover a variety of application domains, such as image recognition,
language translation.

\paragraph{\textbf{Configuration of the parameter search space:}} We considered five
parameters of TensorFlow's threading model for our experiments.  We
described {\interop} and {\intraop} parameters in the Background section
(Section~\ref{section:background}). We provide a brief background of the
other three parameters below. In addition, we describe rationale for selecting
particular ranges of values to tune these parameters. The range
is defined with an upper bound, a lower bound, and a step size value as shown in
Table~\ref{table:parameter_configurations}.

\begin{wraptable}{r}{6.5cm}
\vspace{-0.15in}
\centering
\caption{Tuning parameters and their ranges (min, max, step size)}
\label{table:parameter_configurations}
\begin{footnotesize}
\begin{tabular}{|c|r|r|}
\hline 
\multicolumn{2}{|c|}{\textbf{Parameters}} & \textbf{Range} \\
\hline 
\hline 
\multicolumn{2}{|c|}{{\interop}} & {[}1, 4, 1{]}\\
\hline 
\multicolumn{2}{|c|}{{\intraop}} & {[}1, 56, 1{]}\\
\hline 
\multirow{5}{*}{\batchsize} & NCF & \multirow{2}{*}{{[}64, 256, 64{]}}\\
& SSD-MobileNet & \\ \cline{2-3}
& ResNet50 & \multirow{2}{*}{{[}64, 1024, 64{]}}\\
& Transformer-LT & \\ \cline{2-3}
& BERT & {[}32, 64, 32{]} \\
\hline 
\multicolumn{2}{|c|}{{\kmpblocktime}} & {[}0, 200, 10{]}\\
\hline
\multicolumn{2}{|c|}{{\ompnumthreads}} & {[}1, 56, 1{]}\\
\hline 
\end{tabular}
\end{footnotesize}
\vspace{-0.15in}
\end{wraptable}

\paragraph{{\interop}:} Since this parameter controls the maximum number of concurrent operations
from the data-flow graph, we set this parameter's range from 1 to 4 at step
size of 1. These values were obtained from Intel's recommendation of setting
this parameter based on the number of sockets.

\paragraph{{\intraop}:} Since this parameter controls the maximum number of threads to be used
for operations from the data-flow graphs, we set its range
from 1 to 56 at the step size of 1. This decision was based on Intel's recommendation
of setting {\intraop} based on the number of cores in the system. Intel
Xeon CPUs have per-socket core count of up to 56.

\paragraph{{\batchsize}:} This parameter controls the number of examples provided
to the deep-learning models as input. Setting the value to 1 allows us to obtain
latency of inference, while higher values allow us to obtain
throughput. We note that the batch size is a performance-sensitive parameter ---
a multi-core system could be under-utilized for lower batch size values. Higher
batch sizes thus allow us to explore the saturation points of a system.
Furthermore, some models are computationally-less intensive than others. To
ensure that the target system is satured for all the models, we used different
batch sizes for different models.

\paragraph{{\kmpblocktime}:} This and the next parameter are tunable parameters of
OpenMP runtime library, and they are applicable to Intel-optimized
TensorFlow, since its CPU backend relies on oneDNN library that uses OpenMP library.
{\kmpblocktime} sets the time that a thread should wait
before sleeping after completing the execution of a \emph{parallel region}. Most
deep-learning primitives in oneDNN are implemented using parallel-programming
primitives such as \texttt{parallel for}. The block of code following this
primitive is called ``parallel region'' in OpenMP.
OpenMP tuning guide~\cite{openmp} recommends to set this parameter to 200, but our prior
experiments demonstrated that value of 0 is also sometimes effective.
Consequently, we set the range for this parameter from 0 to 200 with the step size of
10.

\paragraph{{\ompnumthreads}:} This parameter is used to set the maximum number of
threads to use in OpenMP parallel regions. Setting this parameter to a
value higher than 1 enables parallel regions to use multiple cores of
multi-core CPUs concurrently. Intel's guide~\cite{intel-tensorflow-perf-webpage}
recommends setting this parameter to the number of cores in the system. So we
set the range of this parameter to be same as {\intraop}.

\subsection{Results}

We now discuss the results of our tuning experiments. In our experiments, the
models were configured to use 32-bit floating point (FP32) data type.
Additionally, for ResNet50 model, we evaluated it with 8-bit integer (INT8) data
type, which produces a compact model and also reduces its memory footprint.

\begin{figure*}[!t]
\includegraphics[width=\textwidth]{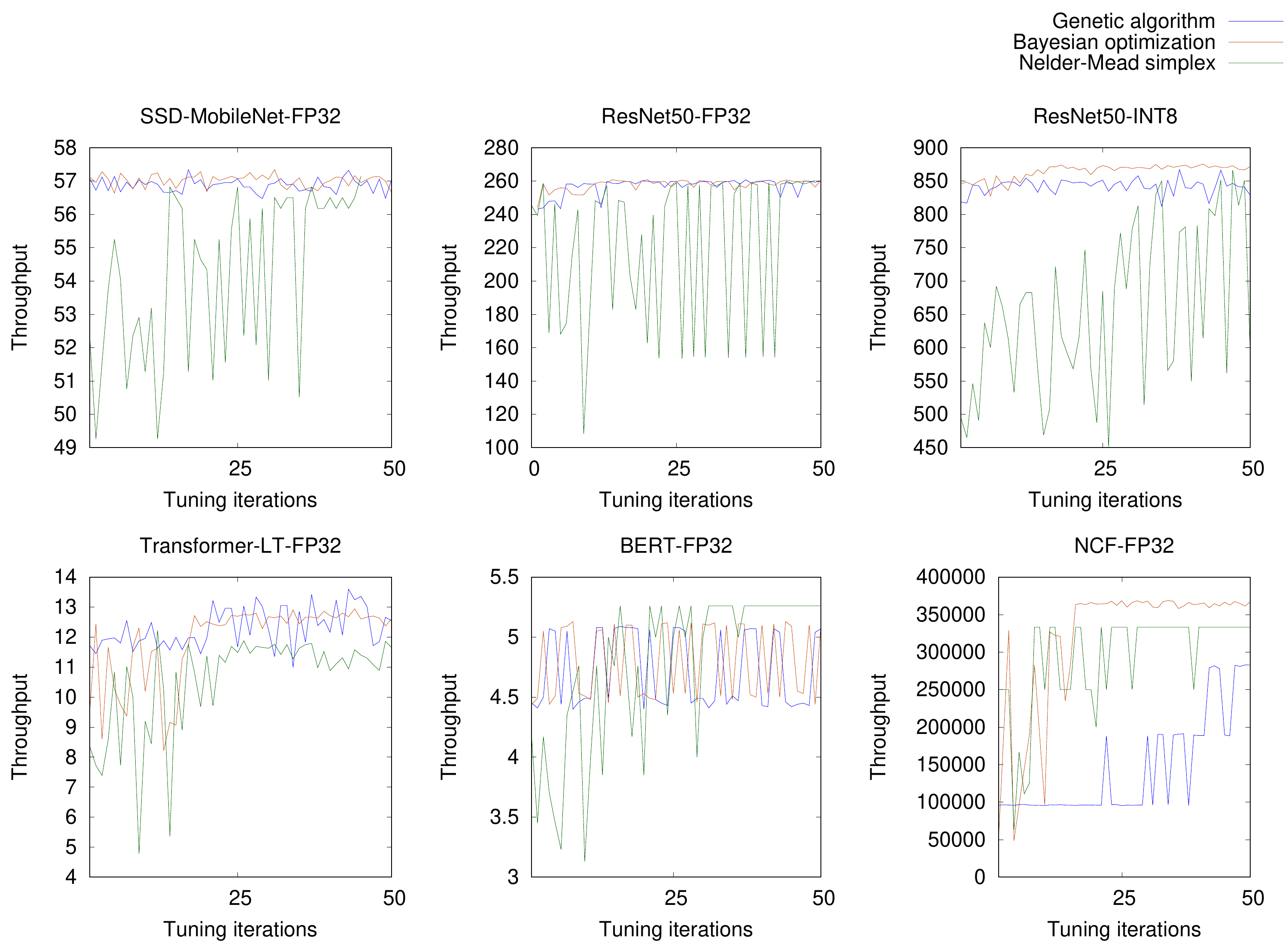}
\caption{Results of auto-tuning TensorFlow's threading model using Bayesian
optimization, genetic algorithm, and Nelder-Mead simplex}
\label{fig:results}
\vspace{-0.1in}
\end{figure*}


Figure~\ref{fig:results} shows the results of tuning the selected deep learning
models using different optimization algorithms. The X axis in the figure represents
tuning iterations (capped at 50), and the Y axis represents the throughput value
(examples/second) --- a higher throughput value represents better performance.
Also, in the figure, green, blue and orange plots represent the performance of
Nelder-Mead simplex, genetic algorithm, and Bayesian optimization,
respectively.

Out of the 6 plots, the top 3 plots for SSD-MobileNet-FP32, ResNet50-FP32, and
ResNet50-Int8 show similar characteristics. Specifically, Bayesian optimization
and genetic algorithm perform similarly and deliver close to peak throughput,
while Nelder-Mead simplex struggles with a considerable variation in the
throughput.  For the bottom 3 plots, namely Transformer-LT-FP32, BERT-FP32, and
NCF-FP32, the optimization algorithms perform differently. Specifically, except
for BERT-FP32, plots for Bayesian optimization and genetic algorithm for other
two models look different. Additionally, while Bayesian optimization delivers
the best performance on tuning NCF-FP32 throughput, it struggles on BERT-FP32,
for which Nelder-Mead simplex delivers the best throughput.  For Transformer-LT
model, genetic algorithm performs better than Nelder-Mead simplex and Bayesian
optimization.

Overall, the results show that no single optimization algorithm consistently
outperforms others in tuning the selected deep learning models.  Nevertheless,
Bayesian optimization demonstrates to be the most competitive overall for the
selection of the models.




\subsection{Comparison of the Optimization Algorithms}

\begin{figure*}[!t]
\includegraphics[width=\linewidth]{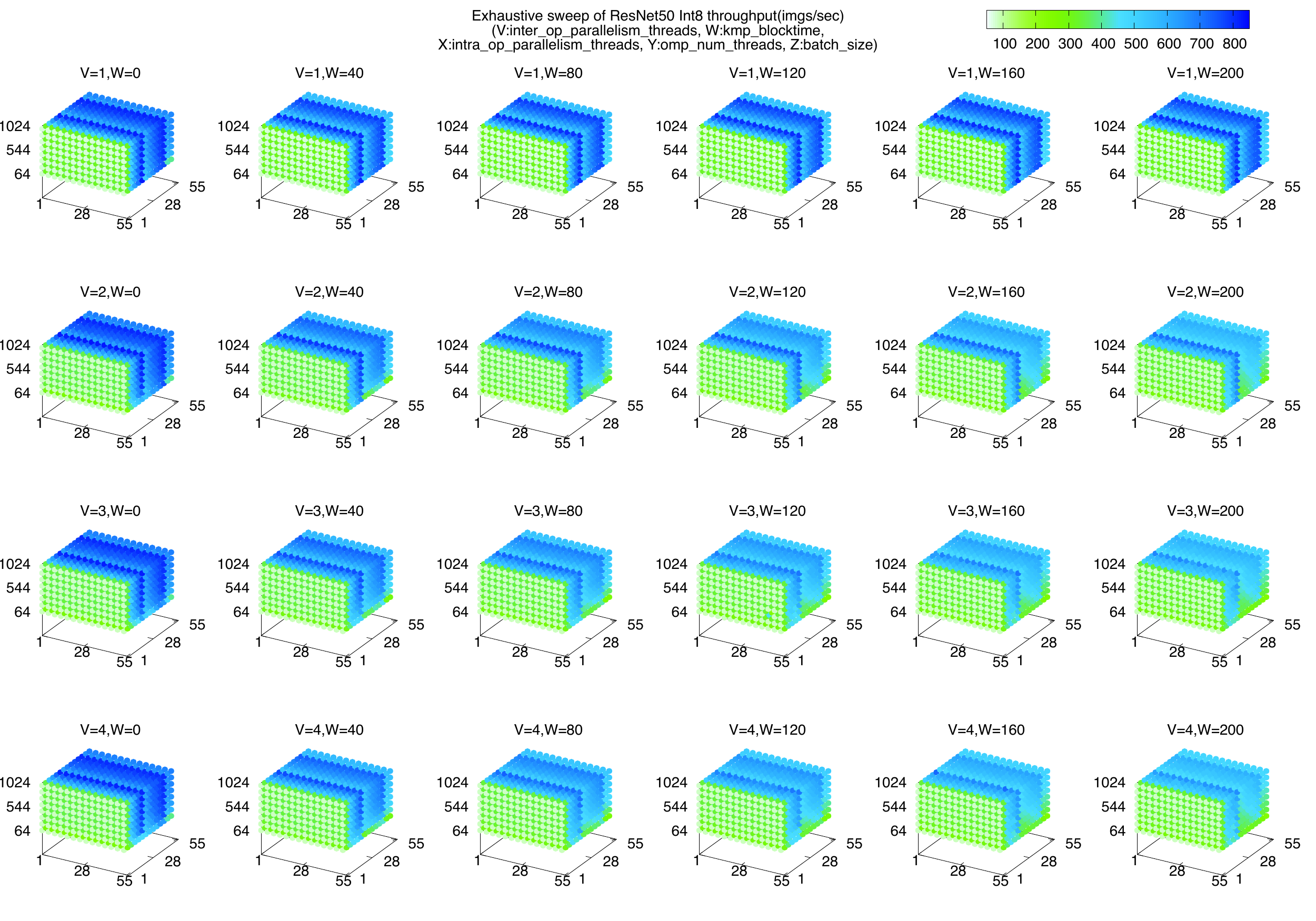}
\caption{Exhaustive sweep of ResNet50-INT8 throughput across all
five parameters. Labels for different axes are in the legend. Colors indicate
different throughput values (dark blue being the highest, and yellow being the
lowest.)}
\label{fig:rn50_exhaustive}
\vspace{-0.1in}
\end{figure*}

\paragraph{\textbf{Exhaustive sweep.}}
In order to understand the effect of different parameters on the performance of
the deep learning models, we performed exhaustive sweep of ResNet50's performance
for INT8 precision across all the five parameters. Figure~\ref{fig:rn50_exhaustive}
shows the throughput of ResNet50 for different parameter values.

We report some salient observations from Figure~\ref{fig:rn50_exhaustive}:
\begin{itemize}

	\item {\kmpblocktime} of 0 delivers better performance than others,
for a given value of {\interop}. 3D plots become lighter in color as
{\kmpblocktime} value increases (from left to right).

  \item All 3D plots have a common pattern: as the value of
{\ompnumthreads} increases (along the Y-axis) the throughput also increases,
suggesting that this parameter has considerable impact on the performance.

	\item Performance does not vary considerably (and noticeably) for different
values of {\intraop} (along the X-axis), suggesting that ResNet50's INT8 model does
not utilize deep learning operators that leverage Eigen threadpool (which relies on
the value of this parameter.)  It also suggests that we can possibly drop this
parameter from the list of tunable parameters to prune the search space.

   \item Batch size (along the Z-axis) has relatively less impact on the
throughput than other parameters, suggesting that it could be dropped from
the list of tunable parameters as well.

\end{itemize}

\paragraph{\textbf{Exploration-exploitation balance.}}

\begin{sidewaysfigure*}
\centering
\begin{subfigure}[t]{0.33\linewidth}
\includegraphics[width=\linewidth]{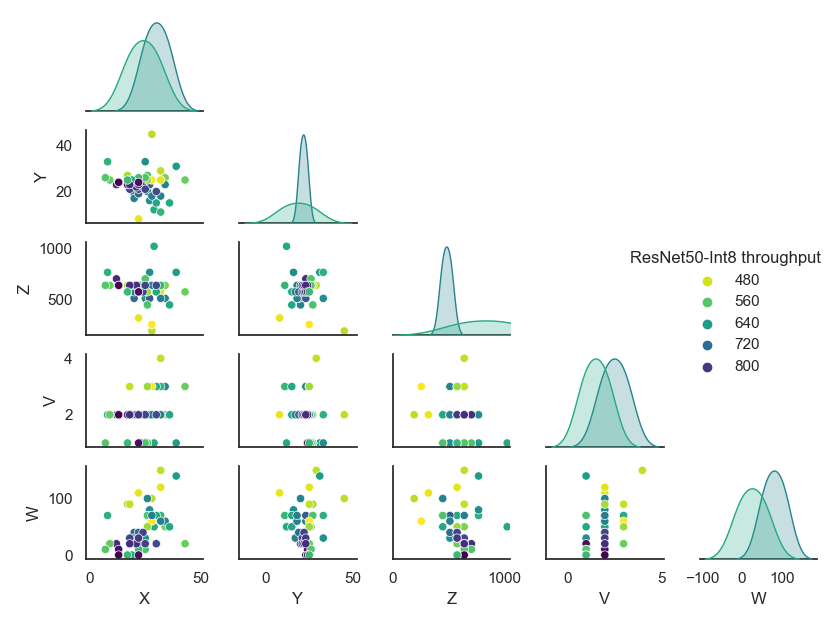}
\caption{Nelder-Mead simplex algorithm}
\label{fig:rn50_tensortuner_pairplot}
\end{subfigure}
\hfill
\begin{subfigure}[t]{0.33\linewidth}
\includegraphics[width=\linewidth]{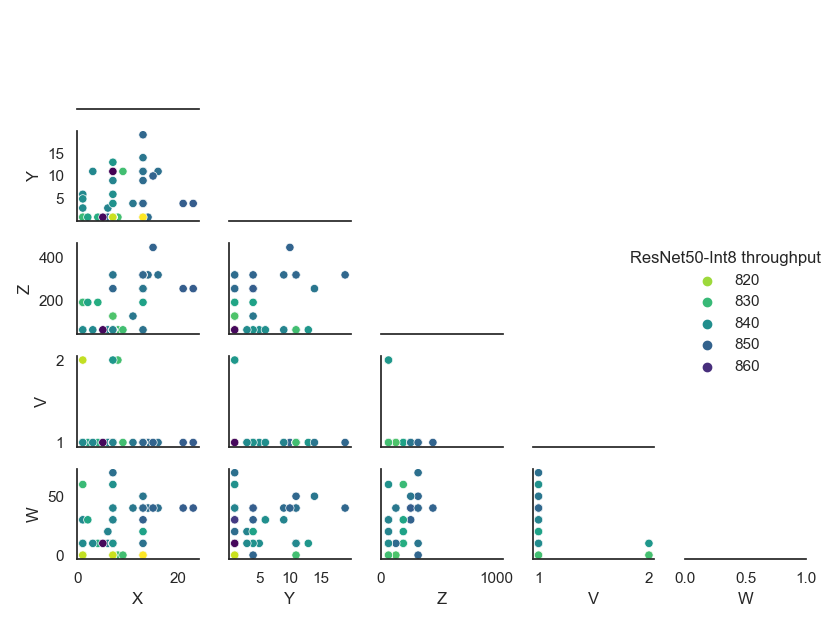}
\caption{Genetic algorithm}
\label{fig:rn50_genetics_pairplot}
\end{subfigure}
\hfill
\begin{subfigure}[t]{0.33\linewidth}
\includegraphics[width=\linewidth]{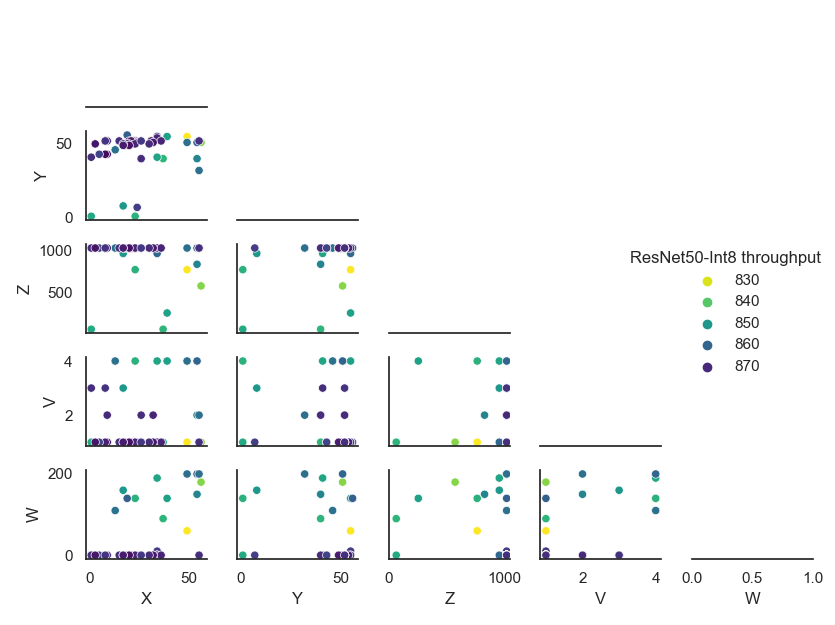}
\caption{Baysian optimization}
\label{fig:rn50_baysian_pairplot}
\end{subfigure}

\centering
\begin{subfigure}[t]{0.33\linewidth}
\includegraphics[width=\linewidth]{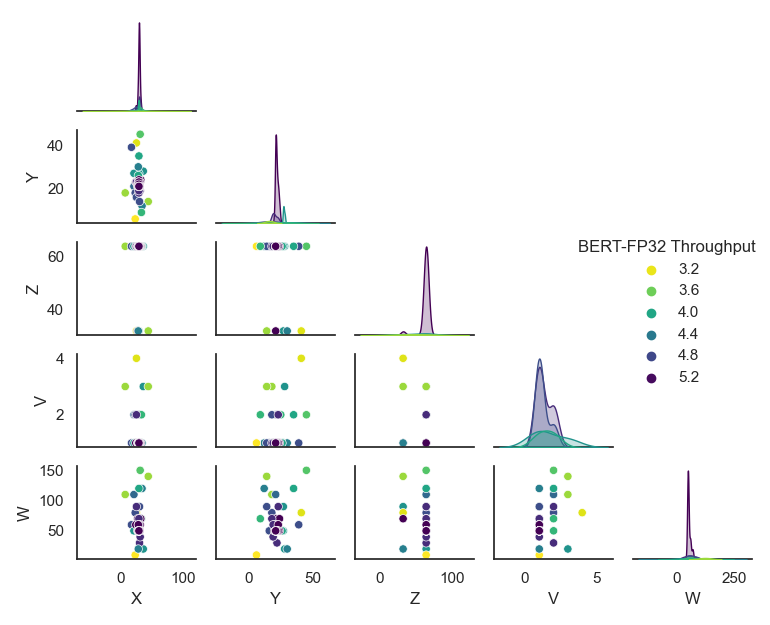}
\caption{Nelder-Mead simplex algorithm}
\label{fig:bert_tensortuner_pairplot}
\end{subfigure}
\hfill
\begin{subfigure}[t]{0.33\linewidth}
\includegraphics[width=\linewidth]{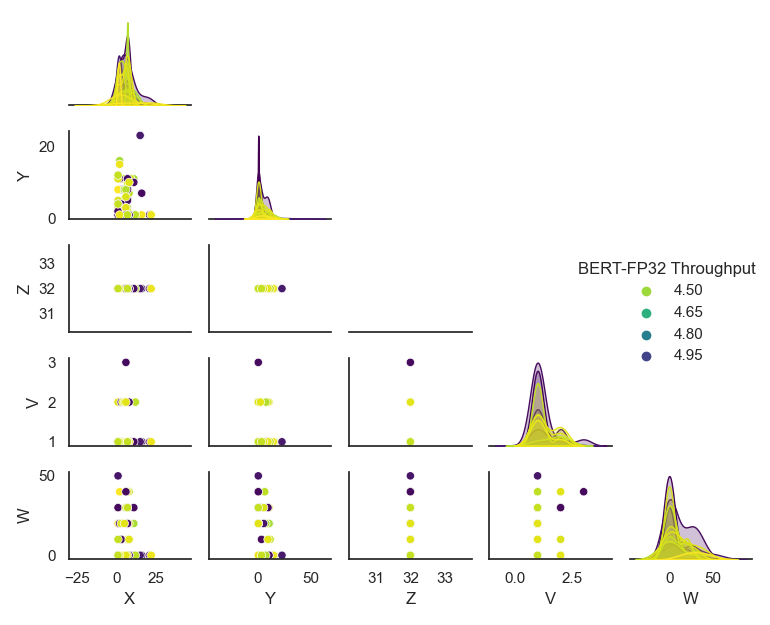}
\caption{Genetic algorithm}
\label{fig:bert_genetics_pairplot}
\end{subfigure}
\hfill
\begin{subfigure}[t]{0.33\linewidth}
\includegraphics[width=\linewidth]{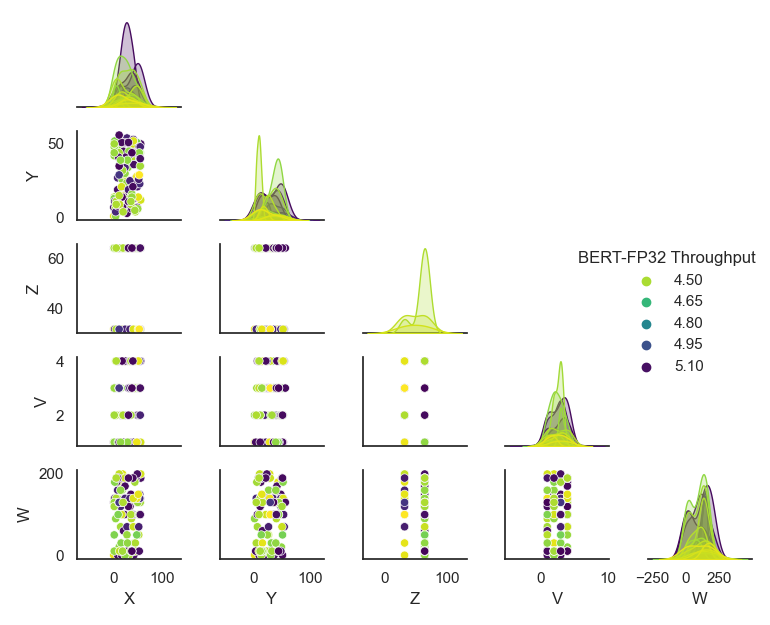}
\caption{Baysian optimization}
\label{fig:bert_baysian_pairplot}
\end{subfigure}
\caption{Pairplots showing configurations sampled by different optimization
algorithms across different parameters and their effect on throughput for
ResNet50-INT8 (top row) and BERT-FP32 (bottom row) models. Parameters:
X={\intraop}, Y={\ompnumthreads},
Z={\batchsize}, V={\interop}, W={\kmpblocktime}}
\label{fig:algo_comparison_pairplot}
\end{sidewaysfigure*}

After obtaining the shape of the performance function for the ResNet50 INT8
model, we compared the effectiveness of all three optimization algorithms in
obtaining optimal configurations. The objective here is to understand the
tradeoff between exploitation and exploration delivered by all three algorithms.
For this purpose, we plotted the ResNet50 INT8 throughput as a function of the
sampled parameter values. Since we have 5 different parameters for tuning, we
convert the 5-dimensional plot into multiple pairplots that represent throughput
as a function of the pairs of the sampled parameter values.
Figure~\ref{fig:rn50_tensortuner_pairplot} shows the ResNet50 INT8 pairplot for
Nelder-Mead simple.  For instance, the plot on the lower left corner between $X$
and $W$ in Figure~\ref{fig:rn50_tensortuner_pairplot} represents the ResNet50
INT8 throughput values in terms of $X$ and $W$ parameter values from the sampled
configuration values.  Darker values represent higher throughput values; lighter
values represent lower throughput values.
Figure~\ref{fig:rn50_baysian_pairplot} shows the pairplot for Bayesian
optimization, and Figure~\ref{fig:rn50_genetics_pairplot} shows the pairplot for
genetic algorithm.

There are a few interesting observations that emerge from these pairplots:
\begin{itemize}

\item Bayesian optimization samples min and max ranges of all the parameters
(seen as squares in pairplots). At the same time, it also samples all the parameters
fairly uniformly, displaying a fair balance between exploitation and exploration.

\item Nelder-Mead simplex algorithm samples clusters of points, indicating that
the algorithm has a higher chance of getting stuck in a local optimum. In other
words, Nelder-Mead simplex devotes more time to exploitation, i.e. local search
in the vicinity of the promising solutions.  Note also that Nelder-Mead simplex
presents another limitation in that it does not sample points along min and max
ranges of some of the parameters (e.g., $Y$ and $Z$),

\item Genetic algorithm, on the other hand, has neither clusters of points nor
samples along min and max ranges (e.g.  parameter $V$, more white spaces in
pairplots), indicating a poor exploration/exploitation balance.

\item Bayesian optimization also has more darker points than Nelder-Mead simplex
algorithm and Genetic algorithm, suggesting that Bayesian optimization delivers
the best balance between exploitation and exploration. This is not surprising,
since this is one of the commonly known advantages of Bayesian optimization.

\end{itemize}

\begin{table*}[t]
\begin{footnotesize}
\begin{tabular}{l||rrrrr||rrrrr}
\hline
\multirow{2}{*}{\textbf{Algorithm}} & \multicolumn{5}{c||}{\textbf{ResNet50
Int8}} & \multicolumn{5}{c}{\textbf{BERT-FP32}} \\ \cline{2-11}
 & X & Y & Z & V & W & X & Y & Z & V & W \\ \hline
Tunable range & [1,56] & [1,56] & [64,1024] & [1,4] & [0,200] & [1,56] & [1,56] &
[32,64] & [1,4] & [0,200] \\ \hline
Nelder-Mead simp- & \multirow{2}{*}{[7,43]} & \multirow{2}{*}{[8,45]} &
  \multirow{2}{*}{[192,1024]} & \multirow{2}{*}{[1,4]} & \multirow{2}{*}{[0,150]}
  & \multirow{2}{*}{[7,44]} & \multirow{2}{*}{[6,45]} & \multirow{2}{*}{[32,64]}
  & \multirow{2}{*}{[1,4]} & \multirow{2}{*}{[10,150]} \\
lex (min, max)& & & & & & & & & & \\ \hline
Genetic algorithm & \multirow{2}{*}{[1,23]} & \multirow{2}{*}{[1,19]} &
  \multirow{2}{*}{[64,448]} & \multirow{2}{*}{[1,2]} & \multirow{2}{*}{[0,70]} &
  \multirow{2}{*}{[1,22]} & \multirow{2}{*}{[1,23]} & \multirow{2}{*}{[32,32]} &
  \multirow{2}{*}{[1,3]} & \multirow{2}{*}{[0,50]} \\
(min,max) & & & & & & & & & & \\ \hline
Bayesian optimi- & \multirow{2}{*}{[1,56]} & \multirow{2}{*}{[1,56]} &
  \multirow{2}{*}{[64,1024]} & \multirow{2}{*}{[1,4]} & \multirow{2}{*}{[0,200]} &
  \multirow{2}{*}{[1,56]} & \multirow{2}{*}{[1,56]} & \multirow{2}{*}{[32,64]} &
  \multirow{2}{*}{[1,4]} & \multirow{2}{*}{[0,200]} \\
zation (min,max) & & & & & & & & & & \\ \hline
Nelder-Mead simp- & \multirow{3}{*}{65} & \multirow{3}{*}{67} &
  \multirow{3}{*}{86} & \multirow{3}{*}{100} & \multirow{3}{*}{75} &
  \multirow{3}{*}{67} & \multirow{3}{*}{70} & \multirow{3}{*}{100} &
  \multirow{3}{*}{100} & \multirow{3}{*}{70} \\
lex sampled & & & & & & & & & &  \\
range(\%) & & & & & & & & & &  \\ \hline
Genetic algorithm & \multirow{2}{*}{40} & \multirow{2}{*}{32} &
  \multirow{2}{*}{40} & \multirow{2}{*}{33} & \multirow{2}{*}{35} &
  \multirow{2}{*}{38} & \multirow{2}{*}{40} & \multirow{2}{*}{50} &
  \multirow{2}{*}{66} & \multirow{2}{*}{25} \\
sampled range(\%) & & & & & & & & & & \\ \hline
Bayesian optimiz- & \multirow{3}{*}{100} & \multirow{3}{*}{100} &
\multirow{3}{*}{100} & \multirow{3}{*}{100} & \multirow{3}{*}{100} &
\multirow{3}{*}{100} & \multirow{3}{*}{100} & \multirow{3}{*}{100} &
\multirow{3}{*}{100} & \multirow{3}{*}{100} \\
ation sampled & & & & & & & & & & \\
range(\%) & & & & & & & & & & \\ \hline
\end{tabular}
\caption{Min/max ranges for different parameters
vs. tunable ranges. Percentages are obtained by
dividing the ranges of the sampled values by the tunable ranges.
Parameters:
X=\texttt{intra\_op\_parallelism\_threads}, Y=\texttt{omp\_num\_threads},
Z=\texttt{batch\_size}, V=\texttt{inter\_op\_parallelism\_threads},
W=\texttt{kmp\_blocktime}}
\label{table:algo_comparison_table}
\end{footnotesize}
\vspace{-0.29in}
\end{table*}

To further understand the exploration-exploitation behavior of all
three algorithms, we obtained their pairplots for the BERT-FP32 model.
Figures~\ref{fig:bert_tensortuner_pairplot}, ~\ref{fig:bert_baysian_pairplot},
~\ref{fig:bert_genetics_pairplot} shows the pairplots for Nelder-Mead simplex
algorithm, Bayesian optimization, and genetic algorithm, respectively.
Observations similar to those of ResNet50-INT8 can be made in BERT's case also.
Specifically, genetic algorithm neither has clusters of points nor samples min/max
ranges, suggesting poor exploration-exploitation strategy. Nelder-Mead
simplex algorithm has clusters of points, similar to that for ResNet50-INT8
model, suggesting that Nelder-Mead simplex algorithm exploits more than
exploring the space (visible from not sampling min/max values of parameter
$V$). However, it still samples better than genetic algorithm in a sense that there is
much less white space in Nelder-Mead simplex's pairplots than in
genetics algorithm's pairplots. Bayesian optimization, on the other hand, has
a cluster of points as well as samples near min/max values, suggesting better
balance between exploration and exploitation.

Overall, these experiments suggest that Bayesian optimization maintains a fair
balance between exploration and exploitation and can sample solutions in
different regions of the search space.  Nelder-Mead simplex algorithm and
genetic algorithm, instead, struggle to maintain this balance, although
Nelder-Mead simplex algorithm in these cases ends up exploring more than genetic
algorithm. This can also been seen from the data in
Table~\ref{table:algo_comparison_table}, which shows the sampled ranges, shown
as (min,max), for different parameters against their tunable ranges for
different optimization algorithms. As can be seen, Bayesian optimization
explores min and max values of all the parameters, both in case of ResNet50-INT8
and BERT-FP32. It explores 100\% of the tunable ranges for all the
parameters for both the models. Genetic algorithm, on the other hand, explores
less than 50\% of the ranges for most of the parameters for both the models,
suggesting that it exploits more than explores.

In summary, we found that no single optimization algorithm can be used to find
the optimal throughput for the selected deep learning workloads. Nevertheless,
Bayesian optimization demonstrates a more robust and reliable behavior and
delivers quality solutions with a limited number of iterations.  Besides, each
algorithm took different time to search the maximum throughput.  For instance,
Nelder-Mead simplex algorithm took a relatively short amount of time for
searching the maximum throughput in BERT-FP32 compared with the others.
Similarly, genetic algorithm took less time to search for the maximum throughput
than Bayesian optimization in SSD-MobileNet-FP32. Moreover, we run our
experiments multiple times, and we observed that the throughput values with both
genetic algorithm and Bayesian optimization are very close in
SSD-MobileNet-FP32, ResNet50-FP32 and BERT-FP32 model.

\section{Related Work}
\label{section:related_work}


In the machine learning domain, auto-tuning is routinely applied to the problem
of hyper-parameter tuning.  Although, there exists a number of commercial and
open-source hyper-parameter tuning systems such as
HyperOpt~\cite{bergstra:2013:hyperopt}, MOE~\cite{moe},
Spearmint~\cite{snoek:2012:spearmint}, AutoWeka~\cite{thornton:2013:autoweka},
SigOpt~\cite{sigopt}, Google's Vizier~\cite{golovin:2017:vizier}, etc., we are
not aware of existing work that systematically analyzes tuning of
performance-sensitive parameters of TensorFlow framework on multi-core CPU
platforms used in data centers. The closest to our work is
TensorTuner~\cite{hasabnis:2018:tensortuner}, which uses Nelder-Mead simplex
algorithm to tune TensorFlow's CPU backend. This work, however, neither
considers other optimization algorithms nor provides insights into the
applicability of Nelder-Mead simplex algorithm to the problem.  Nelder-Mead
simplex is known to be a local optimization algorithm, and global optimization
algorithms such as Bayesian optimization and genetic algorithm exist already.
We seek to provide this comparative evaluation and analysis among different
optimization algorithms on several TensorFlow models, written for a variety of
usecases such as image recognition, language translation, and recommendation
system. TensorTuner, on the other hand, focuses on models used in image
recognition problem. We also consider a larger set of performance-sensitive
tunable parameters, thus expanding the search space of the configurations. Our
work, in this sense, can be considered as an extension of TensorTuner.

Alternative to the auto-tuning based approach considered
in this work, Wang et al.~\cite{wang:2021:tuning-dl-frameworks} develops a
formula to set {\interop} and \\{\intraop} by analyzing the data-flow graphs of
the deep learning models. They develop this formula by analyzing data-flow
graphs of various models and their relationship with the optimal values of these
parameters.  Their approach, however, treats TensorFlow's CPU
backend as a white-box and requires intrusive changes to the TensorFlow
framework. Furthermore, operating as a part of TensorFlow framework, this
approach cannot set other performance-sensitive parameters such as {\batchsize}.
Nevertheless, it is a promising approach in a sense that it can directly set the
values of the parameters by analyzing data-flow graph of a model and thus
eliminate the need of multiple rounds of online or offline tuning.

\section{Conclusion}
\label{section:conclusion}

Overall, our evaluation across a variety of deep learning models for TensorFlow
demonstrates that Bayesian optimization generally maintains good balance
between exploration and exploitation (explores 100\% of the tunable ranges for
all the parameters for ResNet50-INT8 and BERT-FP32 models) and explores most of
the search spaces, if not all. Genetic algorithm, on the other hand, explores
less than 50\% of the ranges for several models and struggles to maintain the
balance between exploration and exploitation.
Nelder-Mead simplex falls in between Bayesian optimization and genetic
algorithm in terms of the exploration-exploitation balance. Nonetheless, we also
found out that no particular optimization algorithm performs the best across all
the models --- Nelder-Mead simplex performs the best on BERT-FP32, while
it lags behind Bayesian optimization and genetic algorithm on the others.

\bibliography{main}
\bibliographystyle{splncs04}

\end{document}